\documentclass[12pt,a4paper]{article}

\usepackage{algorithm}
\usepackage[noend]{algpseudocode}
\usepackage{amsmath}
\usepackage{float}
\usepackage[title]{appendix}
\usepackage{adjustbox,lipsum}
\usepackage{url}
\usepackage{hyperref}
\usepackage{threeparttable}
\usepackage{amssymb}
\usepackage{mathtools}
\usepackage[a4paper,left=2.5cm,right=2.5cm,top=2.5cm,bottom=2.5cm]{geometry}

\begin{document}
\title{Fast Non-Bayesian Poisson Factorization for Implicit-Feedback Recommendations}
\author{David Cortes}
\maketitle

\begin{abstract}
This work explores non-negative low-rank matrix factorization based on regularized Poisson models (PF or "Poisson factorization" for short) for recommender systems with implicit-feedback data. The properties of Poisson likelihood allow a shortcut for very fast computations over zero-valued inputs, and oftentimes results in very sparse factors for both users and items. Compared to HPF (a popular Bayesian formulation of the problem with hierarchical priors), the frequentist optimization-based approach presented here tends to produce better top-N recommendations with significantly shorter fitting times, on top of having sparse solutions.
\end{abstract}

\section{Introduction}

In a typical scenario for recommender systems, a lot of data is available about interactions between users and items, such as users purchasing products or listening to songs, where each user interacts with only a handful of the available items in the catalog.

Typically, recommendation models based on collaborative filtering try to predict entries in the user-item interactions matrix - that is, a matrix in which rows index users, columns index items, and the value at each user-item combination is the number of times that the user has interacted/consumed the item or an explicit rating that the user has given to it - and these predictions are in turn based on minimization of some loss function such as squared difference between the predicted and the real (observed) value, with the idea that items with higher predicted values are better candidates to recommend to the user (see \cite{koren}).

In so-called explicit feedback settings, in which users provide an explicit valuation or rating of each item, such models are usually fit only to the observed (non-missing) entries in the interactions matrix, as this data signals both likes and dislikes of the user, which leads to efficient optimization procedures. However, in so-called implicit-feedback settings, in which there are no explicit ratings but rather only event histories such as songs played by each user, it's not enough for a good model to be fit only to the observed (non-missing) entries, as they don't tend to signal dislikes and there can be pathological cases in which these entries all have the same value (e.g. when the matrix is binary - see \cite{implicit}).

In this case, it's necessary to also consider the unobserved (missing) entries, which is typically done by assigning to them a value of zero, oftentimes reducing the problem to predicting observed vs. unobserved, the latter of which involves orders of magniture more user-item combinations than the former, resulting in a more computationally challenging problem.

Unlike Bernoulli likelihood (log loss) which is typically preferred for this scenario, Poisson likelihood, when using a model that does not exponentiate its parameters, offers a very fast optimization approach for missing entries filled with zeros, since the log-likelihood for them is given by their predicted value only, and in low-rank matrix factorization, the sum of predicted values for all combinations of users and items can be quickly obtained by summing the factorizing matrices column-wise and taking their inner product. This paper proposes exploiting this reduction to fit low-rank matrix factorization models based on counts of interactions instead of only modeling observed vs. unobserved interactions.

This is not the first time that a Poisson model has been proposed for sparse matrix factorization - \cite{gap} also developed this idea, but following a Bayesian approach, while \cite{hpf} improved upon it by adding a hierarchical structure and a faster optimization procedure based on variational inference, with many other works later building upon that base, also relying on variational inference (e.g. \cite{dpf}, \cite{hcpf}). While a hierarchical Bayesian formulation is more expressive (able to capture more complex patterns), fitting such Bayesian models is much slower than conventional optimization techniques applied on regularized frequentist models, and as will be seen, in practice does not end up leading to better local optima.

\section{Low-rank matrix factorization}
Low-rank matrix factorization is one of the most commonly used techniques in collaborative filtering for predicting entries in the user-item interactions matrix based only on observed interactions (\cite{koren}). The idea behind it is to assign to each user $u$ and item $i$ a vector of fixed dimensionality $k$ representing arbitrary features (a.k.a. latent factors or components) $\mathbf{a}_u \in \mathbb{R}^k, \mathbf{b}_i \in \mathbb{R}^k$ (these are the model parameters) in such a way that the value for each entry in the interactions matrix would be approximated by the dot product between the features of the user and the item for that entry, i.e. $x_{ui} \approx \langle \mathbf{a}_u, \mathbf{b}_i \rangle$ or by a transformation of it $x_{ui} \approx f(\langle \mathbf{a}_u, \mathbf{b}_i \rangle)$ - e.g.:
\begin{align}
\min_{\mathbf{A}, \mathbf{B}} \: \lVert I_x(\mathbf{X} - \mathbf{A} \mathbf{B}^T) \lVert
\end{align}
Where $\mathbf{A}_{m, k} = \begin{pmatrix} \mathbf{a}_1, ..., \mathbf{a}_m \end{pmatrix}^T, \mathbf{B}_{n, k} = \begin{pmatrix} \mathbf{b}_1, ..., \mathbf{b}_n \end{pmatrix}^T$, and $I_x$ is the indicator function which is one when the entry $x_{ui}$ is present in the data (item was consumed by the user) and zero when it is missing.

As such model tends to overfit the interactions data, other additional improvements upon it are typically incorporated, such as centering the entries in the matrix, incorporating regularization on the model parameters, and adding user and item biases as additional parameters. The optimization problem is typically solved through the ALS algorithm (Alternating Least-Squares -  see \cite{als}: when one factor matrix is fixed, solving for the other latent factor matrix is a convex optimization problem with a closed-form solution, thus this algorithm alternates between optimizing one or the other matrix until convergence) or through SGD (Stochastic Gradient Descent, see \cite{koren}).

In the implicit-feedback case with missing-as-zero and values consisting of counts (e.g. number of times a user clicked something), biases are  typically left out, entries not centered, and the actual counts are taken as a confidence weight for binary missing-vs-non-missing indicator values instead of being the targets to predict (e.g. \cite{logistic}, \cite{implicit}), resulting in problem formulations such as:

\begin{align}
\min_{\mathbf{A}, \mathbf{B}} \: \lVert g(\mathbf{X}) \odot (I_x - \mathbf{A} \mathbf{B}^T) \lVert^2 + \lambda (\lVert \mathbf{A} \lVert^2 + \lVert \mathbf{B} \lVert^2)
\end{align}

or
$$
\sum_{u} \sum_{i} - g(x_{ui}) \times ll(i_{ui}, f(
	\mathbf{a}_u^T \mathbf{b}_i
))
 + \lambda (\lVert \mathbf{A} \lVert^2 + \lVert \mathbf{B} \lVert^2)
$$
(Where $ll(.)$ is typically the Bernoulli log-likelihood function, $f(.)$ a sigmoid function, and $g(.)$ is a monotonic transformation intended to upweigh non-zero entries such as $1+log(.)$).

This is a more challenging optimization problem, with a matrix $\mathbf{X}$ that usually is too large to even fit in a computer’s memory, but different methods have been devised to solve it or solve variations thereof smartly, such as implicit-ALS (\cite{implicit}) along with techniques to speed it up (\cite{cg}), or BPR (Bayesian Personalized ranking), which tries to sub-sample only some of the missing entries at each update (\cite{bpr}).

\section{Sparse Poisson regression}
A typical probability distribution used for counts data is the Poisson distribution, parameterized by one variable $z > 0$, with probability density function given by $p(y) = z^y \exp(-z) / y!$ . This distribution is limited to non-negative integers and tends to produce asymmetrical and more peaked distributions that are more resemblant of real counts data than symmetric distributions such as the normal distribution. Poisson models in which the $z$ parameter is defined as the sum or dot product of other variables can be fit to observed data by following the maximum-likelihood principle, which translates to maximizing Poisson log-likelihood (the negative of it plus a constant is referred from here on interchangeably as "Poisson loss"), given by:
\begin{align}
ll(z) = - (z - y \log(z) + \log(y!))
\end{align}

Generalized linear models for Poisson regression usually add a link function, taking the form $\mathbf{y} \sim \text{Poisson}(\exp(\mathbf{X} \beta))$, where $\beta$ are the model coefficients (parameters), $\mathbf{X}$ (not to be confused with the matrix in the factorization models) is the matrix of covariates, and $\mathbf{y}$ the observed counts for each observation; but others (e.g. \cite{cmp}) have also tried to perform Poisson regression for all-non-negative covariates without exponentiation (i.e. with an identity link function), constraining the coefficients to be non-negative instead - that is, $\mathbf{y} \sim \text{Poisson}(\mathbf{X} \beta)$, which is the approach that will be followed in this work, since exponentiated numbers would not allow for fast calculation of the sum of all entries in the $\mathbf{X}_{m, n}$ matrix.

As $\log(y!)$ does not depend on the model parameters, it can be left out of the minimization objective. For fitting a Poisson regression with non-negative features of dimensionality $k$ and coefficients without exponentiation to $m$ observations of covariates $\mathbf{A}_{m,k} = \begin{pmatrix} \mathbf{a}_1, ..., \mathbf{a}_m \end{pmatrix}^T$ and counts $\mathbf{b}_m = \begin{pmatrix} b_1, ..., b_m \end{pmatrix}$, the optimization objective (maximum likelihood estimation problem) would look as follows:
\begin{align}
\min_{\mathbf{x} \in \mathbb{R}_{+}^k} \sum_i \mathbf{a}_i^T \mathbf{x} - b_i \log(\mathbf{a}_i^T \mathbf{x})
\end{align}

Note that $\sum_i \mathbf{a}_i^T \mathbf{x}$ can be obtained by first summing $\mathbf{s} = \sum_i \mathbf{a}_i$ and then taking its inner product with $\mathbf{x}$, i.e. $\sum_i \mathbf{a}_i^T \mathbf{x} =  \mathbf{s}^T \mathbf{x}$, something that could not be achieved with the exponentiated version, and when $b_i = 0$, then $b_i \log(\mathbf{a}_i^T \mathbf{x}) = 0$, so for zero-valued entries, maximization of Poisson likelihood translates into minimizing $\mathbf{a}_i^T \mathbf{x}$. As such, the minimization objective can be re-expressed as:
\begin{align}
\min_{\mathbf{x} \in \mathbb{R}_{+}^k} \mathbf{s}^T \mathbf{x} - \sum_{b_i > 0} b_i \log(\mathbf{a}_i^T \mathbf{x})
\end{align}

This is a convex optimization problem, but as others have found out (\cite{cmp}), it cannot be solved through typical methods like L-BFGS-B (\cite{lbfgs}) that rely on assumptions such as Lipschitz continuity or smoothness.

Back to the matrix factorization case, if we adopt Poisson loss (negative likelihood plus a constant) and consider one of the matrices to be fixed, the optimal values for each vector of latent factors in matrices $\mathbf{A}$ and $\mathbf{B}$ are the solution of a Poisson regression problem in which $\mathbf{a}_i$ are the rows of the matrix that was fixed, and $b_i$ are the entries for the corresponding row (for users) or column (for items) in the interactions matrix $\mathbf{X}$. From the formula above, it can be seen that Poisson loss can be calculated without ever iterating through the non-zero values (their contribution is obtained through $\mathbf{s}$), which is very convenient and efficient in the implicit-feedback case as most of the entries will indeed be zero.

\section{Solving Poisson regression}
Finding the minimizers for the sparse Poisson regression with identity link function presents some challenges:
\begin{itemize}
	\item Coefficients need to be restricted to be non-negative.
	\item The end result of the dot products for all observed user-item interactions must be restricted to be strictly greater than zero.
	\item The function to optimize does not have a Lipschitz gradient.
	\item The Hessian of the function is not guaranteed to be positive-semi-definite.
\end{itemize}

Most typically used methods such as L-BFGS-B (\cite{lbfgs}) try to take iterative approximate Newton steps satisfying strong Wolfe conditions, but as others have found, these are oftentimes not possible to satisfy with non-smooth functions (see e.g. \cite{nonsmooth}), oftentimes leading to earlier termination with suboptimal results. Not satisfying either strong or weak Wolfe conditions can lead to problems such as making a quasi-Newton approximation non-positive-semi-definite, which makes their application to sparse Poisson regression problematic.

What's more, the typical approach for dealing with bound constraints in second-order methods is to find at each step the point that minimizes the function value of the second order approximation subject to the bound constraints, but such an approach can lead to solutions that do not satisfy the strict inequality of dot products for each observed user-item combination in the original problem here. A potential solution is to add further inequality constraints for each observed user-item pair, but this makes the problem much larger and adds more challenges in the optimization producedure. Another potential solution is to add intercepts so that there would always be some way of overturning an exact-zero prediction, but this approach turned out to produce much worse solutions when evaluated under ranking metrics.

As such, a potential approach is to use gradient-based optimization methods that restrict their steps to driving only one additional variable to its bound at each time.

One could alternatively think of splitting the function into:
$$
f(\mathbf{x}) = g(\mathbf{x}) + h(\mathbf{x})
$$
With
$$
g(\mathbf{x}) = -\sum_{b_i > 0} b_i \log(\mathbf{a}_i^T \mathbf{x}),
\:\:\:\:\:\:\:\:\:\:\:\:
h(\mathbf{x}) = \mathbf{s}^T \mathbf{x}
$$

And try to apply methods based on forward-backward splitting (e.g. \cite{boyd}) which are much faster for this problem, but this approach presents numerical instability issues, and in practice, using small steps for this problem tended to produce solutions in which all predictions end up being very similar for every user (non-personalized), so the forward-backward splitting approach was not pursued further.

Some approaches that are particularly well suited for this problem are the conjugate gradient methods from \cite{nncg} and \cite{tncg} (the implementation used here was adapted from SciPy\footnote{\url{https://docs.scipy.org/doc/scipy/reference/optimize.minimize-tnc.html}}, which is in turn adapted from \cite{scipy_tncg}), both of which take steps that limit the active set (variables taking a value of zero, which is their lower bound) to one new variable at a time and do not require positive-definiteness of (real) Hessians to work correctly, as they work based only on evaluation of gradients. The gradient for the objective function, assuming $L_2$ regularization, is given by:
$$
\nabla f(\mathbf{x}) = - \sum_{b_i > 0} \frac{b_i}{\mathbf{a}_i^T \mathbf{x}} \mathbf{a}_i + \mathbf{s} + 2 \lambda \mathbf{x}
$$

($L_1$ regularization could also be easily incorporated by adding to $\mathbf{s}$)

The method from \cite{nncg} has faster steps, while \cite{tncg} makes more progress towards minimization with each step taken. Both of them were evaluated here, and both were able to find the minimizers for sparse Poisson regression in most cases, but with some caveats: \cite{tncg} initialized with a bad starting point that is close to the optimum, can oftentimes get stuck due to meeting its termination criteria without having reached the actual optimum, requiring a careful initialization; while \cite{nncg} can present numerical precision issues with adjustment of step sizes and might end up requiring a lot of steps in some cases.

\section{Poisson Matrix Factorization}
Back to the case of low-rank matrix factorization, a regularized factorization with Poisson likelihood can be expressed as the following problem:
$$
\min_{\mathbf{A} \in \mathbb{R}_{+}^{m,k}, \mathbf{B} \in \mathbb{R}_{+}^{n,k}}
\left( \sum_{u=1}^{m} \sum_{i=1}^{n}
\mathbf{a}_u^T \mathbf{b}_{i} -
x_{ui} \log (\mathbf{a}_u^T \mathbf{b}_{i})
\right)
+ \lambda ( || \mathbf{A} ||_2^2 + || \mathbf{B} ||_2^2 )
$$

From the previous sections, we know that the following re-expression is equivalent:
$$
\min_{\mathbf{A} \in \mathbb{R}_{+}^{m,k}, \mathbf{B} \in \mathbb{R}_{+}^{n,k}}
\mathbf{s}_A \mathbf{s}_B^T
- \sum_{x_{ui} > 0} x_{ui} \log (\mathbf{a}_u^T \mathbf{b}_{i})
+ \lambda ( || \mathbf{A} ||_2^2 + || \mathbf{B} ||_2^2 )
$$

Where $\mathbf{s}_A$ and $\mathbf{s}_B$ are row vectors consisting of the column-wise sums of $\mathbf{A}$ and $\mathbf{B}$, respectively.

One way to find a local minimum to this problem is by following a block coordinate descent producedure based on alternating minimization similarly to ALS:
\begin{algorithm}[H]
\caption{Alternating Minimization for Poisson Factorization}\label{Alternating Minimization for Poisson Factorization}
\hspace*{\algorithmicindent} \textbf{Inputs} Sparse matrix $\mathbf{X} \in \mathbb{R}_{+}^{m, n}$, regularization parameter $\lambda$, number of iterations $T$, initial non-negative values for $\mathbf{A}_{m,k}, \mathbf{B}_{m,k}$  \\
\hspace*{\algorithmicindent} \textbf{Outputs} Optimized latent factors or components $\mathbf{A}^*, \mathbf{B}^*$
\begin{algorithmic}[1]
\For {$1 .. T$}
	\State Calculate $\mathbf{s}_B = \sum_i^n \mathbf{b}_i$
	\For {$u = 1 .. m$}
		\State Update
		$
		\mathbf{a}_u \leftarrow
		\text{argmin}_{\mathbf{a}_u}
		\mathbf{s}_B \mathbf{a}_u
		- \sum_{x_{ui} > 0}
		x_{ui} \log (\mathbf{a}_u^T \mathbf{b}_{i})
		+ \lambda || \mathbf{a}_u ||_2^2
		$
	\EndFor
	
	\State Calculate $\mathbf{s}_A = \sum_i^m \mathbf{a}_i$
	\For {$i = 1 .. n$}
		\State Update
		$
		\mathbf{b}_i \leftarrow
		\text{argmin}_{\mathbf{b}_i}
		\mathbf{s}_A \mathbf{b}_i
		- \sum_{x_{ui} > 0}
		x_{ui} \log (\mathbf{b}_i^T \mathbf{a}_{u})
		+ \lambda || \mathbf{b}_i ||_2^2
		$
	\EndFor
\EndFor
\Return $\mathbf{A}, \mathbf{B}$
\end{algorithmic}
\end{algorithm}

The updates for each $\mathbf{a}_u$ and $\mathbf{b}_i$ in turn can be solved by some conjugate gradient method such as \cite{nncg} or \cite{tncg} (but note that not just \textit{any} CG-type method would do, see e.g. \cite{bccg}), and do not need to be solved exactly every time as the procedure will iterate again over them later with now better values for the other factors that are held as fixed.

The matrices can be initialized to random small numbers drawn from a uniform or gamma distribution. One initialization that proved particularly good was that used by \cite{hpf} in its original source code\footnote{\url{https://github.com/premgopalan/hgaprec}}: $0.3 + \epsilon$, with $\epsilon \sim \text{Uniform}(0, 10^{-2})$.

The original ALS technique with squared errors suggested that convergence would be achieved after around 12 iterations and that more iterations would only increase the quality of the obtained solutions, as would adding extra factors, but in the case of Poisson factorization, there are some additional aspects to consider:
\begin{itemize}
\item The non-negativity constraints and presence of a summed term make the optimal solution for a given user or item vector at a given round have potentially many values being exactly zero.
\item Once many values in the matrix that is fixed at a given round are at exactly zero, solving for the other matrix might not necessarily change the values from the previous iteration, making it more likely to get stuck in a given local optimum and in some cases impossible to escape it if only one matrix is being optimized at a time.
\item If step sizes are restricted by each individual factor, adding extra factors can significantly shrink the possible step sizes and increase the amount of steps required to reach convergence, despite the calculations of gradients and steps scaling linearly with $k$, but at the same time, this could make it less likely to get stuck in a bad local optimum due to the values that are exactly zero in the other matrix.
\item Adding more factors may significantly change the dynamics of which variables are driven to the bounds at each iteration and which are taken off from the bounds, which makes converging to a local optimum harder.
\end{itemize}

In the ALS scenario with unobserved entries from the user-item interactions matrix ignored, regularization is typically scaled for each user and item based on the number of observed entries, whereas here all entries are considered in the objective function so it makes more sense to use the same regularization for all users and items. Some quick experimentation on public datasets suggests that potentially good $\lambda$ candidates for large datasets would be between $10^2$ to $10^7$, with $10^3$ being a good default value for $k=40$ and with the method from \cite{tncg} being better at solving problems with low regularization compared to \cite{nncg}. Some $L_1$ regularization was also experimented with but it turned out to provide significantly worse results than $L_2$ regularization.

When using the method from \cite{tncg}, convergence is typically achieved by 10 iterations already, with further iterations not changing a single value for any factor, while \cite{nncg} with the problem solved inexactly (and it did require more iterations to solve each subproblem exactly) required around 20-30 iterations to reach convergence when limiting each subproblem to 10-30 updates/steps (with the actual optimum sometimes requiring more than 100 steps to reach, while \cite{tncg} typically requires less than 50 steps).

Both methods presented some numerical issues, with \cite{nncg} in some cases failing to find an optimum and \cite{tncg} in many cases benefiting from a starting point far from the optimum that doesn't reuse previous solutions - in this sense, a potentially good starting point is assigning the same small constant to all factors being solved for, such as $10^{-3}$.

When predicting the values for unobserved user-item combinations (those that have a value of zero in the $\mathbf{X}$ to which the model was fit), the predicted values can end up being exactly zero, which makes their Poisson likelihood undefined - as such, it is not possible to keep a separate validation set on which metrics are to be calculated in order to decide when to stop the optimization procedure (as done in \cite{hpf}), and calculating model quality metrics on a held-out test set based on Poisson likelihood (also done in \cite{hpf}) is also likely to fail; but otherwise this aspect of having predicted values being exactly zero can be helpful when items for a given user need to be ranked.

Potential termination criteria to look at for outer iterations are the norm of the difference w.r.t. previous iteration or simply running for a pre-defined maximum number of iterations without evaluating any termination criteria along the way.

\section{Experiments}
The Poisson factorization model for collaborative filtering described here (its implementation was made open source and freely available\footnote{\url{https://github.com/david-cortes/poismf}}) was compared against its Bayesian counterpart HPF (\cite{hpf}) fit through through variational-inference-based coordinate ascent, to the BPR (Bayesian Personalized Ranking, see \cite{bpr}) method with sub-sampled negative entries, and to the implicit-ALS method with weighted squared errors on binarized entries, using public datasets for recommendation systems on implicit-feedback data and evaluated through ranking-based metrics on a random held-out test sample.

As a baseline, non-personalized recommendations were also compared against, which were produced by an intercepts-only version of implicit-ALS\footnote{\url{https://github.com/david-cortes/cmfrec}} - that is, each item getting a score calculated by a weighted sum of its interactions divided by the number of available users plus a regularization term, and each user given the same recommendation as obtained by ranking items in descending order of this score.

The datasets used were the RetailRocket\footnote{\url{https://www.kaggle.com/retailrocket/ecommerce-dataset}}, Last.FM-360K (\cite{lastfm}) and EchoNest MillionSong (\cite{millionsong}) datasets:
\begin{table}[H]
\caption {Dataset Descriptions}
\begin{adjustbox}{max width=\textwidth}{\centering
\begin{tabular}{|l|c|c|c|c|}
 \hline

\textbf{Dataset} & \textbf{\# Users} & \textbf{\# Items} & \textbf{\# Entries} & \textbf{Percent non-zeros} \\ \hline
\textbf{RetailRocket} & 1,407,580 & 235,061 & 2,145,179 & 0.00065\% \\ \hline
\textbf{Last.FM-360K} & 358,868 & 160,113 & 17,535,655 & 0.03\% \\ \hline
\textbf{EchoNest} & 1,019,318 & 384,546 & 48,373,586 & 0.012\% \\ \hline

\hline
\end{tabular}}\end{adjustbox}
\end{table}

The RetailRocket dataset contains events "click", "add to basket", and "purchase" in an e-commerce shop. In order to set a count for each user-item pair, the events were given the following values: click = +1, add to basket = +3, purchase = +3, with the final value for a given user-item pair being the sum of the event values for it. The Last.FM-360K and EchoNest MillionSong datasets contain counts of the number of times a song was played by a sample of users in online music listening services - all of the entries in these two datasets were used with the values they came, without applying any transformation, but note that, for methods like implicit-ALS, better results can be obtained in these two datasets by a transformation that would shrink the weights compared to the raw counts.

All the methods were compared using their default/suggested hyperparameters in publicly available implementations\footnote{\url{https://github.com/benfred/implicit/}}\footnote{\url{https://github.com/david-cortes/hpfrec}}, with some modifications:
\begin{itemize}
\item Each model was fitted using 40, 70, and 100 latent factors/components, with the remainder of the hyperparameters kept at their default values. Due to memory constraints, HPF was limited to only 40 factors in the largest dataset.
\item Although some of the papers introducing each model suggested evaluating different termination criteria, each model was run for a fixed number of outer iterations (100 for HPF and for BPR, 15 for ALS), without any time spent on evaluation of termination criteria for outer iterations, but with inner iterations (e.g. conjugate gradient steps) still terminating according to their own criteria.
\item PF (the model proposed here) was run using different inner CG solvers, running for 10 outer iterations when using \cite{tncg} and for 30 outer iterations when using \cite{nncg} (plus limiting each inner iteration to 5 steps per round), using an $L_2$ regularization of $10^3$ for \cite{tncg} and $10^4$ for \cite{nncg}, and running with both a constant initialization at each time and a reuse of previous solutions when using the CG solver from \cite{tncg}. These are referred hereafter as:
	\begin{itemize}
	\item Variant 1: using the CG method from \cite{nncg}.
	\item Variant 2: using the CG method from \cite{tncg}, reusing previous solutions as starting point at each iteration.
	\item Variant 3: using the CG method from \cite{tncg}, but re-initializing the factors at each update to the same small value of $10^{-3}$.
	\end{itemize}
\item All factors were calculated in single precision (32-bit floating point numbers).
\end{itemize}

Models were evaluated by selecting a random sample of 10,000 users as test set, from which 30\% of their consumed items (calculated for each user individually) were set as a test set, considering only cases in which the test set would contain at least two observed items. Then models were fit on the remainder of the data, and items that did not appear as consumed in the training data to which the model was fit were ranked using the obtained factors for each user and item, considering the held-out items as positive examples and the remainder as negative examples. The following metrics were calculated (individually for each user and then averaged across the 10,000 test users):
\begin{itemize}
\item P@10 ("precision-at-10"), indicating the number of positive items present among the top-10 ranked items (with consumed items from the training data excluded).
\item MAP ("mean average precision"), measuring the average precision (true positive rate) at each recall level (percent of identified positives) under the (full) ranking produced by the model.
\item NDCG@10 ("net discounted cumulative gain at 10"), which is a metric that takes into account not only whether an item was consumed or not, but also the value associated with it (e.g. number of times that a given user played each song).
\end{itemize}
The implementation of the training-test split and evaluation metrics used here was also made open source and freely available\footnote{\url{https://github.com/david-cortes/recometrics}}.

A perhaps more realistic evaluation methodology would be to exclude the test users altogether from the optimization procedure and then calculate latent factors/components for them from the 70\% of non-held-out data using the fitted item factors alone, but unfortunately, many of the software implementations compared against did not support such calculations.

Times were measured on an AMD Ryzen 7 2700 CPU with 8 cores (16 threads) running at 3.2Ghz, and all the software used here was compiled with GCC version 10.3 with flags "-O3 -march=native -flto -fno-math-errno -fno-trapping-math". These timings measure time taken to fit each model to the \textbf{full} data (\textbf{without} performing a train-test split).

Results are as follows:
\begin{table}[H]
\caption {Results on RetailRocket Dataset}
\begin{adjustbox}{max width=\textwidth}{\centering
\begin{tabular}{|l|c|c|c|c|c|}
 \hline

\textbf{Model} & \textbf{k} & \textbf{P@10} & \textbf{MAP} & \textbf{NDCG@10} & \textbf{Time (s)} \\ \hline
\textbf{Non-personalized} & - & 0.0013 & 0.0024 & 0.0028 & 0.028 \\ \hline \hline
\textbf{PF - variant 1} & 40 & 0.0078 & 0.0169 & 0.0206 & 36.37 \\ \hline
 \textbf{PF - variant 2} & 40 & 0.0008 & 0.0013 & 0.0017 & 25.76 \\ \hline
\textbf{PF - variant 3} & 40 & 0.0027 & 0.0065 & 0.0083 & 36.68 \\ \hline
\textbf{HPF} & 40 & 0.0036 & 0.0070 & 0.0092 & 91 \\ \hline
\textbf{iALS} & 40 & 0.0128 & 0.0273 & 0.0358 & 20.64 \\ \hline
\textbf{BPR} & 40 & \textbf{0.0199} & \textbf{0.0440} & \textbf{0.0568} & \textbf{17.65} \\ \hline \hline

\textbf{PF - variant 1} & 70 & 0.0105 & 0.0221 & 0.0269 & 41.99 \\ \hline
\textbf{PF - variant 2} & 70 & 0.0006 & 0.0013 & 0.0016 & 47.42 \\ \hline
\textbf{PF - variant 3} & 70 & 0.0028 & 0.0063 & 0.0078 & 74 \\ \hline
\textbf{HPF} & 70 & 0.0049 & 0.0103 & 0.0131 & 152 \\ \hline
\textbf{iALS} & 70 & 0.0165 & 0.0361 & 0.0471 & 25.88 \\ \hline
\textbf{BPR} & 70 & \textbf{0.0199} & \textbf{0.0432} & \textbf{0.0556} & \textbf{25.31} \\ \hline \hline

\textbf{PF - variant 1} & 100 & 0.0129 & 0.0284 & 0.0359 & 46.42 \\ \hline
\textbf{PF - variant 2} & 100 & 0.0009 & 0.0014 & 0.0019 & 76 \\ \hline
\textbf{PF - variant 3} & 100 & 0.0040 & 0.0090 & 0.0114 & 119 \\ \hline
\textbf{HPF} & 100 & 0.0055 & 0.0127 & 0.0159 & 215 \\ \hline
\textbf{iALS} & 100 & 0.0196 & 0.0420 & 0.0554 & \textbf{31.76} \\ \hline
\textbf{BPR} & 100 & \textbf{0.0202} & \textbf{0.0434} & \textbf{0.0565} & 35.38 \\ \hline

\hline
\end{tabular}}\end{adjustbox}
\end{table}

\begin{table}[H]
\caption {Results on Last.FM-360K Dataset}
\begin{adjustbox}{max width=\textwidth}{\centering
\begin{tabular}{|l|c|c|c|c|c|}
 \hline

\textbf{Model} & \textbf{k} & \textbf{P@10} & \textbf{MAP} & \textbf{NDCG@10} & \textbf{Time (s)} \\ \hline
\textbf{Non-personalized} & - & 0.0463 & 0.0296 & 0.0453 & 0.126 \\ \hline \hline
\textbf{PF - variant 1} & 40 & 0.0769 & 0.0479 & 0.0616 & 192 \\ \hline
\textbf{PF - variant 2} & 40 & 0.1227 & 0.0808 & 0.1241 & 149 \\ \hline
\textbf{PF - variant 3} & 40 & 0.1213 & 0.0795 & 0.1198 & 455 \\ \hline
\textbf{HPF} & 40 & 0.1162 & 0.0769 & 0.1169 & 501 \\ \hline
\textbf{iALS} & 40 & \textbf{0.1655} & \textbf{0.1171} & \textbf{0.1647} & \textbf{20.05} \\ \hline
\textbf{BPR} & 40 & 0.0823 & 0.0511 & 0.0862 & 120 \\ \hline \hline
\textbf{PF - variant 1} & 70 & 0.0012 & 0.0180 & 0.0007 & 186 \\ \hline
\textbf{PF - variant 2} & 70 & 0.1353 & 0.0907 & 0.1390 & 221 \\ \hline
\textbf{PF - variant 3} & 70 & 0.1383 & 0.0923 & 0.1407 & 848 \\ \hline
\textbf{HPF} & 70 & 0.1199 & 0.0796 & 0.1219 & 857 \\ \hline
\textbf{iALS} & 70 & \textbf{0.1745} & \textbf{0.1252} & \textbf{0.1721} & \textbf{26.33} \\ \hline
\textbf{BPR} & 70 & 0.0842 & 0.0520 & 0.0893 & 160 \\ \hline \hline
\textbf{PF - variant 1} & 100 & 0.1053 & 0.0697 & 0.1035 & 209 \\ \hline
\textbf{PF - variant 2} & 100 & 0.1460 & 0.0984 & 0.1504 & 313 \\ \hline
\textbf{PF - variant 3} & 100 & 0.1436 & 0.0966 & 0.1472 & 1321 \\ \hline
\textbf{HPF} & 100 & 0.1259 & 0.0822 & 0.1268 & 1230 \\ \hline
\textbf{iALS} & 100 & \textbf{0.1790} & \textbf{0.1289} & \textbf{0.1768} & \textbf{32.78} \\ \hline
\textbf{BPR} & 100 & 0.0840 & 0.0513 & 0.0866 & 215 \\ \hline
\hline
\end{tabular}}\end{adjustbox}
\end{table}

\begin{table}[H]
\caption {Results on EchoNest MillionSong Dataset}
\begin{adjustbox}{max width=\textwidth}{\centering
\begin{tabular}{|l|c|c|c|c|c|}
 \hline

\textbf{Model} & \textbf{k} & \textbf{P@10} & \textbf{MAP} & \textbf{NDCG@10} & \textbf{Time (s)} \\ \hline
\textbf{Non-personalized} & - & 0.0248 & 0.0200 & 0.0302 & 0.371 \\ \hline \hline
\textbf{PF - variant 1} & 40 & 0.0013 & 0.0082 & 0.0014 & 608 \\ \hline
\textbf{PF - variant 2} & 40 & 0.0377 & 0.0294 & 0.0457 & 423 \\ \hline
\textbf{PF - variant 3} & 40 & 0.0375 & 0.0307 & 0.0462 & 1151 \\ \hline
\textbf{HPF} & 40 & 0.0374 & 0.0313 & 0.0471 & 1448 \\ \hline
\textbf{iALS} & 40 & \textbf{0.0695} & \textbf{0.0532} & \textbf{0.0835} & \textbf{53.12} \\ \hline
\textbf{BPR} & 40 & 0.0420 & 0.0384 & 0.0469 & 431 \\ \hline \hline
\textbf{PF - variant 1} & 70 & 0.0017 & 0.0111 & 0.0020 & 408 \\ \hline
\textbf{PF - variant 2} & 70 & 0.0392 & 0.0321 & 0.0494 & 669 \\ \hline
\textbf{PF - variant 3} & 70 & 0.0433 & 0.0351 & 0.0524 & 2195 \\ \hline
\textbf{iALS} & 70 & \textbf{0.0802} & \textbf{0.0614} & \textbf{0.0956} & \textbf{68} \\ \hline
\textbf{BPR} & 70 & 0.0454 & 0.0408 & 0.0503 & 580 \\ \hline \hline
\textbf{PF - variant 1} & 100 & 0.0389 & 0.0316 & 0.0467 & 424 \\ \hline
\textbf{PF - variant 2} & 100 & 0.0410 & 0.0328 & 0.0510 & 1014 \\ \hline
\textbf{PF - variant 3} & 100 & 0.0448 & 0.0354 & 0.0524 & 3481 \\ \hline
\textbf{iALS} & 100 & \textbf{0.0887} & \textbf{0.0668} & \textbf{0.1036} & \textbf{82} \\ \hline
\textbf{BPR} & 100 & 0.0481 & 0.0424 & 0.0529 & 766 \\ \hline
\hline
\end{tabular}}\end{adjustbox}
\end{table}

\section{Discussion}
The frequentist optimization-based approach to Poisson factorization presented here (PF) managed to consistently obtain similar or better results than its Bayesian counterpart while taking less time (and requiring much less memory) in two of the datasets experimented with, and better results on the remaining dataset under one of the solvers but not the other. The CG method from \cite{nncg} presented numerical instabilities that made it achieve very poor results when the number of factors is small (this was due to some predictions becoming exactly zero) but otherwise good results, while the CG method from \cite{tncg} tended to achieve stable results in less total ellapsed time, particularly when reusing previous solutions as starting points for the next iteration.

Although \cite{hpf} experimented with some of the same datasets used here and reported an improvement in both ranking quality metrics and model fitting times compared to a matrix factorization that minimizes squared error, the comparison there was against an explicit-feedback variant (ignoring items not consumed by users) applied on implicit-feedback data, using a stochastic optimization method which is quite sensitive to optimization parameters, and the timings measured single-threaded executions of the algorithms - if the comparison is instead performed against a squared-error variant that takes missing entries into account, the quality of the rankings for top-N items in these datasets seems consistently better under the squared-error variant, and a more careful software reimplementation of both methods exploiting parallelization and block coordinate descent methods makes the squared-error version significantly faster than the Poisson versions (either frequentist or Bayesian).

Even though the software implementations of Poisson factorization developed here have some room for speed improvements - for example, the HPF implementation used a digamma function (taken from SciPy\footnote{\url{https://docs.scipy.org/doc/scipy/reference/generated/scipy.special.psi.html}}) designed for arbitrary 64-bit floats while always receiving non-negative numbers and using the result to calculate a 32-bit sum, and the PF line search was implemented by re-computing full dot-products instead of keeping the search direction dot products separately and shrinking them at every trial before summing them to the current values - the possible room for potential speed ups is rather small, and the large difference in speed between the squared-error factorization and the Poisson factorizations would still look similar under more optimized software implementations.

Both Poisson factorization approaches (frequentist and Bayesian) were nevertheless competitive against BPR (Bayesian personalized ranking), which seemed to only perform well in the most sparse dataset used here, and contrary to the results reported in \cite{bpr}, did not perform any better than implicit-ALS in most cases.

\section{Sparse factors}
A very interesting aspect of the Poisson factorization presented here is that, depending on the solver used, the optimal factors for each user and item might end up being very sparse - that is, a majority of their entries having values of exactly zero - which can be exploited in order to compute rankings faster and makes similarity searches and clustering easier, faster, and more intuitive, as each user and item can be though of as having affinity for only a very limited subset of factors, which also opens the doors for more potential approximations on different tasks.

Even though solutions may look similar when evaluated under ranking metrics, the sparsity patterns that they produce differ a lot under each optimization approach, with the CG method from \cite{nncg} in particular (variant 1) achieving much sparser solutions, and a blank start in \cite{tncg} (variant 3) oftentimes preventing solutions from becoming sparse if the maximum steps per update are not enough. Although the solutions have a large number of zero-valued entries in each vector, there were no only-zeros columns, so this model cannot be said to be automatically selecting $k$, at least not with the values experimented with (40, 70, 100).

\begin{table}[H]
\caption {Sparsity patterns for models in Last.FM-360K dataset}
\begin{adjustbox}{max width=\textwidth}{\centering
\begin{tabular}{|l|c|c|c|c|}
 \hline

\textbf{Metric} & \textbf{Variant} & \textbf{k} & \textbf{User Factors} & \textbf{Item Factors} \\ \hline
\textbf{Percent zero-valued} & Variant 1 & 40 & 75.45\% & 94.30\% \\ \hline
\textbf{Percent zero-valued} & Variant 2 & 40 & 41.45\% & 61.07\% \\ \hline
\textbf{Percent zero-valued} & Variant 3 & 40 & 0.00\% & 0.00\% \\ \hline

\textbf{Mean non-zeros per vector} & Variant 1 & 40 & 9.82 & 2.28 \\ \hline
\textbf{Mean non-zeros per vector} & Variant 2 & 40 & 23.42 & 15.57 \\ \hline
\textbf{Mean non-zeros per vector} & Variant 3 & 40 & 40 & 40 \\ \hline
\textbf{Median non-zeros per vector} & Variant 1 & 40 & 9 & 1 \\ \hline
\textbf{Median non-zeros per vector} & Variant 2 & 40 & 23 & 16 \\ \hline
\textbf{Median non-zeros per vector} & Variant 3 & 40 & 40 & 40 \\ \hline \hline

\textbf{Percent zero-valued} & Variant 1 & 70 & 72.10\% & 94.88\% \\ \hline
\textbf{Percent zero-valued} & Variant 2 & 70 & 43.63\% & 67.76\% \\ \hline
\textbf{Percent zero-valued} & Variant 3 & 70 & 0.00\% & 0.00\% \\ \hline

\textbf{Mean non-zeros per vector} & Variant 1 & 70 & 19.53 & 3.58 \\ \hline
\textbf{Mean non-zeros per vector} & Variant 2 & 70 & 39.46 & 22.57 \\ \hline
\textbf{Mean non-zeros per vector} & Variant 3 & 70 & 70 & 70 \\ \hline
\textbf{Median non-zeros per vector} & Variant 1 & 70 & 19 & 2 \\ \hline
\textbf{Median non-zeros per vector} & Variant 2 & 70 & 39 & 25 \\ \hline
\textbf{Median non-zeros per vector} & Variant 3 & 70 & 70 & 70 \\ \hline \hline

\textbf{Percent zero-valued} & Variant 1 & 100 & 73.33\% & 91.56\% \\ \hline
\textbf{Percent zero-valued} & Variant 2 & 100 & 44.53\% & 71.73\% \\ \hline
\textbf{Percent zero-valued} & Variant 3 & 100 & 0.00\% & 0.00\% \\ \hline

\textbf{Mean non-zeros per vector} & Variant 1 & 100 & 26.67 & 8.44 \\ \hline
\textbf{Mean non-zeros per vector} & Variant 2 & 100 & 55.47 & 28.27 \\ \hline
\textbf{Mean non-zeros per vector} & Variant 2 & 100 & 100 & 100 \\ \hline
\textbf{Median non-zeros per vector} & Variant 1 & 100 & 25 & 8 \\ \hline
\textbf{Median non-zeros per vector} & Variant 2 & 100 & 56 & 32 \\ \hline
\textbf{Median non-zeros per vector} & Variant 2 & 100 & 100 & 100 \\ \hline

\hline
\end{tabular}}\end{adjustbox}
\end{table}

\begin{table}[H]
\caption {Sparsity patterns for models in EchoNest MillionSong dataset}
\begin{adjustbox}{max width=\textwidth}{\centering
\begin{tabular}{|l|c|c|c|c|}
 \hline

\textbf{Metric} & \textbf{Variant} & \textbf{k} & \textbf{User Factors} & \textbf{Item Factors} \\ \hline
\textbf{Percent zero-valued} & Variant 1 & 40 & 81.65\% & 78.07\% \\ \hline
\textbf{Percent zero-valued} & Variant 2 & 40 & 38.20\% & 68.19\%
 \\ \hline
\textbf{Percent zero-valued} & Variant 3 & 40 & 0.00\% & 0.00\%
 \\ \hline
\textbf{Mean non-zeros per vector} & Variant 1 & 40 & 7.34 & 8.77 \\ \hline
\textbf{Mean non-zeros per vector} & Variant 2 & 40 & 24.72 & 12.72 \\ \hline
\textbf{Mean non-zeros per vector} & Variant 3 & 40 & 40 & 40 \\ \hline
\textbf{Median non-zeros per vector} & Variant 1 & 40 & 6 & 6 \\ \hline
\textbf{Median non-zeros per vector} & Variant 2 & 40 & 24 & 12 \\ \hline
\textbf{Median non-zeros per vector} & Variant 3 & 40 & 40 & 40 \\ \hline \hline

\textbf{Percent zero-valued} & Variant 1 & 70 & 79.02\% & 84.04\% \\ \hline
\textbf{Percent zero-valued} & Variant 2 & 70 & 41.23\% & 69.73\% \\ \hline
\textbf{Percent zero-valued} & Variant 3 & 70 & 0.00\% & 0.00\% \\ \hline

\textbf{Mean non-zeros per} vector & Variant 1 & 70 & 14.68 & 11.17 \\ \hline
\textbf{Mean non-zeros per vector} & Variant 2 & 70 & 41.14 & 21.19 \\ \hline
\textbf{Mean non-zeros per vector} & Variant 3 & 70 & 70 & 70 \\ \hline
\textbf{Median non-zeros per vector} & Variant 1 & 70 & 14 & 8 \\ \hline
\textbf{Median non-zeros per vector} & Variant 2 & 70 & 41 & 23 \\ \hline
\textbf{Median non-zeros per vector} & Variant 3 & 70 & 70 & 70 \\ \hline \hline

\textbf{Percent zero-valued} & Variant 1 & 100 & 83.33\% & 92.38\% \\ \hline
\textbf{Percent zero-valued} & Variant 2 & 100 & 42.25\% & 70.17\% \\ \hline
\textbf{Percent zero-valued} & Variant 3 & 100 & 0.00\% & 0.00\% \\ \hline

\textbf{Mean non-zeros per vector} & Variant 1 & 100 & 16.67 & 7.62 \\ \hline
\textbf{Mean non-zeros per vector} & Variant 2 & 100 & 57.75 & 29.83 \\ \hline
\textbf{Mean non-zeros per vector} & Variant 3 & 100 & 100 & 100 \\ \hline
\textbf{Median non-zeros per vector} & Variant 1 & 100 & 14 & 7 \\ \hline
\textbf{Median non-zeros per vector} & Variant 2 & 100 & 57 & 31 \\ \hline
\textbf{Median non-zeros per vector} & Variant 3 & 100 & 100 & 100 \\ \hline

\hline
\end{tabular}}\end{adjustbox}
\end{table}

(Note that "variant 3" can also achieve sparse solutions comparable to "variant 1" if the maximum number of steps per update is increased much further than what was suggested in the user manual of its implementation, at the expense of even longer fitting times)

In contrast, the hierarchical prior imposed by HPF in \cite{hpf} makes the optimal solutions for every factor strictly greater than zero, thus preventing any potential sparsity and making its application less attractive even if ranking quality results were similar.

\section{Conclussions}
This work presented a frequentist optimization-based approach to low-rank Poisson factorization of sparse counts matrices, which proved to be significantly faster and achieve better results for recommender systems applications than a Bayesian formulation of the problem with hierarchical priors, despite in theory being less expressive.

While \cite{hpf} reported promising results for Poisson matrix factorization applied to recommender systems with implicit-feedback data compared to squared-error factorization, a more careful comparison that takes into account unobserved interactions in a squared-error variant suggests that the squared-error factorization does a better job at producing top-N recommendations and takes much less time to calculate its factors.

Nevertheless, the Poisson factorization presented here was able to produce very sparse factors, with some matrices having $>90\%$ non-zeros, which makes it an attractive option as this sparsity pattern can simplify and speed up many tasks such as ranking, clustering and similarity searches, which is a property that is not shared by the Bayesian formulation.

\bibliographystyle{plain}
\bibliography{poisrec}

\end{document}